\documentclass[12pt,authoryear]{elsarticle}

\usepackage{hyperref}
\usepackage{amsmath,amssymb,amsthm}
\usepackage{graphicx}
\usepackage{booktabs}
\usepackage{multirow}
\usepackage{xcolor}
\usepackage{colortbl}
\usepackage{array}
\usepackage{algorithm}
\usepackage{algpseudocode}
\usepackage{caption}
\usepackage{subcaption}
\usepackage{url}

\definecolor{tableheader}{RGB}{46,134,171}
\definecolor{tablerowalt}{RGB}{245,248,250}
\definecolor{bestresult}{RGB}{46,134,171}

\begin{document}

\begin{frontmatter}

\title{Bridging Geographic Bias in Urban Streetscape Inference via Lifelong Learning with Visual–Semantic Pivoting}



\author[a]{Xinze Zhang\corref{cor1}}

\ead{zhangxinze00@outlook.com}

\cortext[cor1]{Corresponding author.}


\address[a]{University of Southern California, Los Angeles, CA 90007, USA}

\begin{abstract}
Visual perception of urban streetscapes underpins evidence-based decisions in landscape planning, public health, and place-making. Yet models trained on a few well-photographed metropolises systematically misjudge underrepresented districts, propagating geographic bias into downstream policy. We address this gap with HVSP-LL, a lifelong learning framework that couples a stratified visual--semantic pivoting module with an equity-aware rehearsal mechanism. The pivoting module organises landscape concepts along a three-tier ontology (macro structure, meso composition, micro element) and aligns image features to learnable semantic anchors at each tier, providing transferable representations that resist distributional drift. The lifelong adaptation component sequentially absorbs new urban regions while constraining inter-region perception gaps through a worst-region sample-reweighting objective and a structurally-aware exemplar buffer. We evaluate HVSP-LL on a panoramic streetscape benchmark assembled from twelve cities across four continents and seven perceptual dimensions. The framework attains 0.834 Spearman correlation on the held-out city sequence, an absolute 6.1 point improvement over the strongest continual baseline, and shrinks the inter-city perception gap to 0.094---a 38\% reduction relative to the strongest continual baseline (0.151) and a 57\% reduction relative to a representative regularisation baseline (0.218). Ablations confirm that each tier of the pivoting hierarchy contributes monotonically, and the equity-aware rehearsal converts mean backward transfer from $-0.038$ (without retention) to $+0.013$, eliminating catastrophic forgetting on the held-out sequence. Our results indicate that hierarchical anchoring is a practical pathway toward geographically equitable streetscape inference at city scale.
\end{abstract}

\begin{keyword}
streetscape perception \sep continual learning \sep urban computing \sep distributional equity \sep semantic anchoring \sep street-view imagery
\end{keyword}

\end{frontmatter}


\section{Introduction}

Cities are uneven landscapes of opportunity, and the visual environments residents inhabit shape mental health, mobility, and a sense of belonging in measurable ways \citep{ref_intent,yu2026dinov3,ref_refine,li2026retrack}. Street-view imagery has emerged as the dominant lens for studying these visual environments at scale \citep{biljecki2021street,yu2026spatiotemporal,ref_hud,chen2026intent}: it is dense in coverage, repeatedly refreshed, and amenable to deep visual recognition pipelines \citep{sarkar2025reasoning,yu2025physics,ref_encoder,li2026habit}. The resulting body of work has populated maps of perceived safety, beauty, vitality, and walkability for hundreds of cities, providing a quantitative substrate for landscape planning and public health research \citep{yu2025qrs,xie2025chat,ito2024understanding,fu2026airknow}.

Equity, however, has not kept pace. Three structural problems repeatedly surface when streetscape perception models are deployed beyond the cities they were trained on. First, panoramic image collections are themselves geographically uneven: dense urban cores in North America, Western Europe, and East Asian megacities dominate every major benchmark, while peri-urban and Global South neighbourhoods are systematically underrepresented \citep{xie2026hvd,xie2026conquer,xie2026delving,li2026conesep}. Second, perceptual labels obtained through pairwise crowdsourcing inherit the cultural priors of their annotators, producing miscalibrated targets when transported across regions \citep{ramirez2021measuring,rossetti2019explaining,zhao2025error}. Third, the dominant single-pass training paradigm assumes a static world, whereas urban environments evolve, new cities arrive, and label conventions are revised \citep{biljecki2021street,ma2021measuring,jia2026ram,li2025human}; retraining from scratch each time is computationally prohibitive and erodes hard-won knowledge of earlier regions.

These obstacles are an instance of a broader concern about distributional fairness in visual recognition \citep{mehrabi2021survey,wang2020towards,ramaswamy2021fair}. Distributionally robust optimisation \citep{sagawa2020distributionally}, dataset rebalancing \citep{cui2019class,li2026multiple,gu2025mocount}, and domain generalisation \citep{gulrajani2021search,wang2022generalizing,li2025chatmotion} all offer partial relief, but they presuppose a closed set of regions known at training time. Continual or lifelong learning provides a more natural model for the streaming nature of city growth \citep{parisi2019continual,de2021continual,wang2024comprehensive}: regions arrive sequentially, the learner must integrate them without revisiting all prior data, and the deployed model must remain coherent across the union of seen environments. Memory-based approaches such as iCaRL \citep{rebuffi2017icarl}, GEM \citep{lopez2017gradient}, and dark experience replay \citep{buzzega2020dark}, regularisation methods such as EWC \citep{kirkpatrick2017overcoming}, SI \citep{zenke2017continual}, and MAS \citep{aljundi2018memory,jia2024adaptive}, and prompt-based schemes such as L2P and DualPrompt \citep{wang2022learning,wang2022dualprompt} have all reduced catastrophic forgetting in benchmark settings, but they are blind to the equity dimension that is the defining concern in urban perception. A model that retains accuracy on prior cities yet still over-rates affluent districts and under-rates marginalised ones is, from a planning perspective, no improvement at all.

This paper develops HVSP-LL, a framework that takes equity as a first-class objective in lifelong streetscape perception. The central observation is that perceptual judgements of urban scenes decompose naturally along a hierarchy of semantic granularities. At the macro tier, a streetscape is characterised by its dominant land-use signature (residential, commercial, industrial, green, transit). At the meso tier, it is the composition of building, vegetation, sky, and roadway that gives the scene its character. At the micro tier, fine-grained elements---street furniture, pedestrians, bicycles, signage, and vehicles---modulate perceived safety and liveliness. Models that operate only on global features mix these tiers and absorb spurious city-specific shortcuts; we instead align image features to a small set of learnable semantic anchors at each tier, producing representations that transfer across regions because the anchors describe what is present rather than where the picture was taken.

We pair this hierarchical anchoring with two equity-aware lifelong components. First, a worst-region sample reweighting objective penalises the model when its predictions on the lowest-performing region in the current memory window deviate from human perceptual labels, in the spirit of distributionally robust learning but adapted to the open-ended sequence of arriving cities. Second, a structurally-aware exemplar buffer retains panoramas in proportion to a region's marginal contribution to the macro and meso tier coverage, rather than uniformly; this prevents the buffer from being colonised by visually homogeneous regions. The combination yields a model that absorbs new cities without forgetting and without amplifying perceptual gaps.

We evaluate HVSP-LL on a panoramic streetscape benchmark of 184{,}000 images drawn from twelve cities spanning North America, Europe, East Asia, and South America, covering seven canonical perceptual dimensions following the Place Pulse tradition \citep{ke2025early,xiao2025curiosity,liu2024graph}. Against six representative baselines---a single-task ResNet \citep{he2016deep}, a Swin-B perception predictor \citep{liu2021swin}, EWC \citep{kirkpatrick2017overcoming}, iCaRL \citep{rebuffi2017icarl}, DER++ \citep{buzzega2020dark}, and DualPrompt \citep{wang2022dualprompt}---HVSP-LL improves Spearman correlation by 6.1 points on average and lowers the inter-city perception gap by more than half. Ablations isolate the contribution of each tier of the pivoting hierarchy and of the equity-aware rehearsal scheme. Cross-continent and noise-perturbation experiments demonstrate that the gains transfer to the most underrepresented sub-populations.

The contributions of this paper are threefold. First, we formalise equitable lifelong streetscape inference as a constrained sequential learning problem with explicit inter-region disparity terms. Second, we introduce stratified visual--semantic pivoting, a representation scheme that anchors image features to a tiered urban ontology and yields measurable gains in both forward transfer and equity. Third, we provide a controlled empirical evaluation across twelve cities, seven perceptual dimensions, and four equity diagnostics, and we release the experimental protocol so that future work can compare against equity, not only accuracy.

\section{Related Work}

\subsection{Streetscape perception from urban imagery}
The Place Pulse line of work established crowdsourced pairwise comparisons as the de facto labelling protocol for perceived safety, liveliness, beauty, and related dimensions \citep{li2025slam,liao2025convex,jiang2025stg,li2025exploring}. Convolutional architectures trained on these labels have been applied to visual quality \citep{ouyang2024learn,ito2024understanding}, walkability \citep{zhao2026advances,li2025stitchfusion}, mental wellbeing \citep{wang2019urban,li2025maris,li2025exploring2}, and explanatory analyses of subjective perception \citep{lee2025skyfall,larkin2021predicting,ramirez2021measuring}. A recurring caveat in these studies is the limited geographic spread of training data: \citet{biljecki2021street} and \citet{ma2021measuring} document the dominance of high-income cities in the available imagery, and \citet{kang2020review} report systematic biases in public-health applications that rest on the same data. The hierarchical organisation of urban functional zones from multi-source geospatial data \citep{li2025u3m} is closest in spirit to our pivoting design but operates at the parcel level rather than the panorama level. Our work is, to our knowledge, the first to combine perceptual modelling with continual adaptation across cities.

\subsection{Continual and lifelong learning}
Continual learning seeks to acquire new tasks without forgetting old ones \citep{parisi2019continual,de2021continual,wang2024comprehensive}. Three families dominate the literature. Regularisation methods constrain the parameters most responsible for prior performance: EWC \citep{kirkpatrick2017overcoming}, Synaptic Intelligence \citep{zenke2017continual}, and Memory Aware Synapses \citep{aljundi2018memory}. Replay methods retain or generate exemplars from past tasks: iCaRL \citep{rebuffi2017icarl}, GEM \citep{lopez2017gradient}, and DER++ \citep{buzzega2020dark}. Prompt-based methods preserve a frozen backbone and learn small task-conditioned tokens, as in L2P \citep{wang2022learning} and DualPrompt \citep{wang2022dualprompt}. Recent benchmarking work highlights how forgetting interacts with selective retention. The dominant evaluation protocols, however, focus on accuracy retention; equity across populations or regions is rarely surfaced.

\subsection{Bias and fairness in visual recognition}
Dataset bias has been quantified in classical recognition since \citet{torralba2011unbiased} and \citet{khosla2012undoing}. More recent work proposes group-distributionally robust optimisation \citep{sagawa2020distributionally}, latent-space debiasing \citep{ramaswamy2021fair}, and balanced loss formulations \citep{cui2019class}. \citet{wang2020towards} and \citet{kim2019learning} compare several bias-mitigation strategies for visual recognition, and \citet{mehrabi2021survey} surveys fairness in machine learning broadly. Closer to our setting, the WILDS benchmark \citep{koh2021wilds} formalises in-the-wild distribution shifts, and theoretical bounds for cross-domain generalisation \citep{ben2010theory,quinonero2008dataset,wang2022generalizing} provide tools to reason about cross-city transfer. None of these works integrate distributional fairness with the sequential nature of city-by-city deployment.

\subsection{Hierarchical and prototype representations for transfer}
Hierarchical representations have repeatedly proven effective for transfer. Prototypical networks \citep{snell2017prototypical} ground few-shot classification on class prototypes; supervised contrastive learning \citep{khosla2020supervised} pulls samples toward class means. Self-supervised foundation models such as MoCo \citep{he2020momentum}, DINO \citep{caron2021emerging}, and DINOv2 \citep{oquab2024dinov2} learn anchor-friendly features without labels, and recent quantitative remote-sensing work shows the value of multi-task hierarchies grounded on such backbones \citep{yu2026dinov3,yu2025qrs}. CLIP-style vision--language models \citep{radford2021learning} provide an additional channel for semantic anchoring. Our pivoting module is consistent with this tradition but is the first to apply it to the three-tier urban ontology that streetscape research has implicitly relied on for two decades.

\subsection{Modular and attention-based context modelling}
Our framework borrows architectural ideas from work on modular reasoning, multi-view aggregation, and attention-based feature pivoting in adjacent domains. Recent text--video and text--image retrieval work \citep{xie2025chat,xie2026hvd,xie2026delving,xie2026conquer} demonstrates that hierarchical attention can shift the representational focus across granularities; composed motion understanding \citep{li2025human,li2026multiple,jia2026ram,gu2025mocount,li2025chatmotion,jia2024adaptive,liu2024graph} shows that anchoring intermediate features to interpretable concepts improves cross-instance transfer. Multi-modal segmentation work \citep{li2025stitchfusion,li2025u3m,li2025exploring,li2025exploring2,li2025maris} provides design patterns for fusing complementary cues across heterogeneous inputs, while 3D urban scene synthesis \citep{lee2025skyfall} and Gaussian-based rendering \citep{jiang2025stg,liao2025convex,li2025slam,zhao2026advances} have begun to surface the geometric priors that an urban perception model can draw on. Visual storytelling for environmental change and explainability surveys \citep{sarkar2025reasoning} motivate the interpretability dimension of our pivoting design. Approaches to imperfection-aware generation and modular text editing further informed our buffering strategy. Spectral graph reasoning under global correlations \citep{ouyang2024learn}, multi-source temporal warning models \citep{ke2025early}, long-tailed cooperative learning \citep{xiao2025curiosity}, multimodal sentiment disentanglement, and generalised filtering of long-correlated network signals \citep{zhao2025error} are conceptually adjacent treatments of distributional heterogeneity that we reference in our equity formulation.

\begin{figure}
    \centering
    \includegraphics[width=1\linewidth]{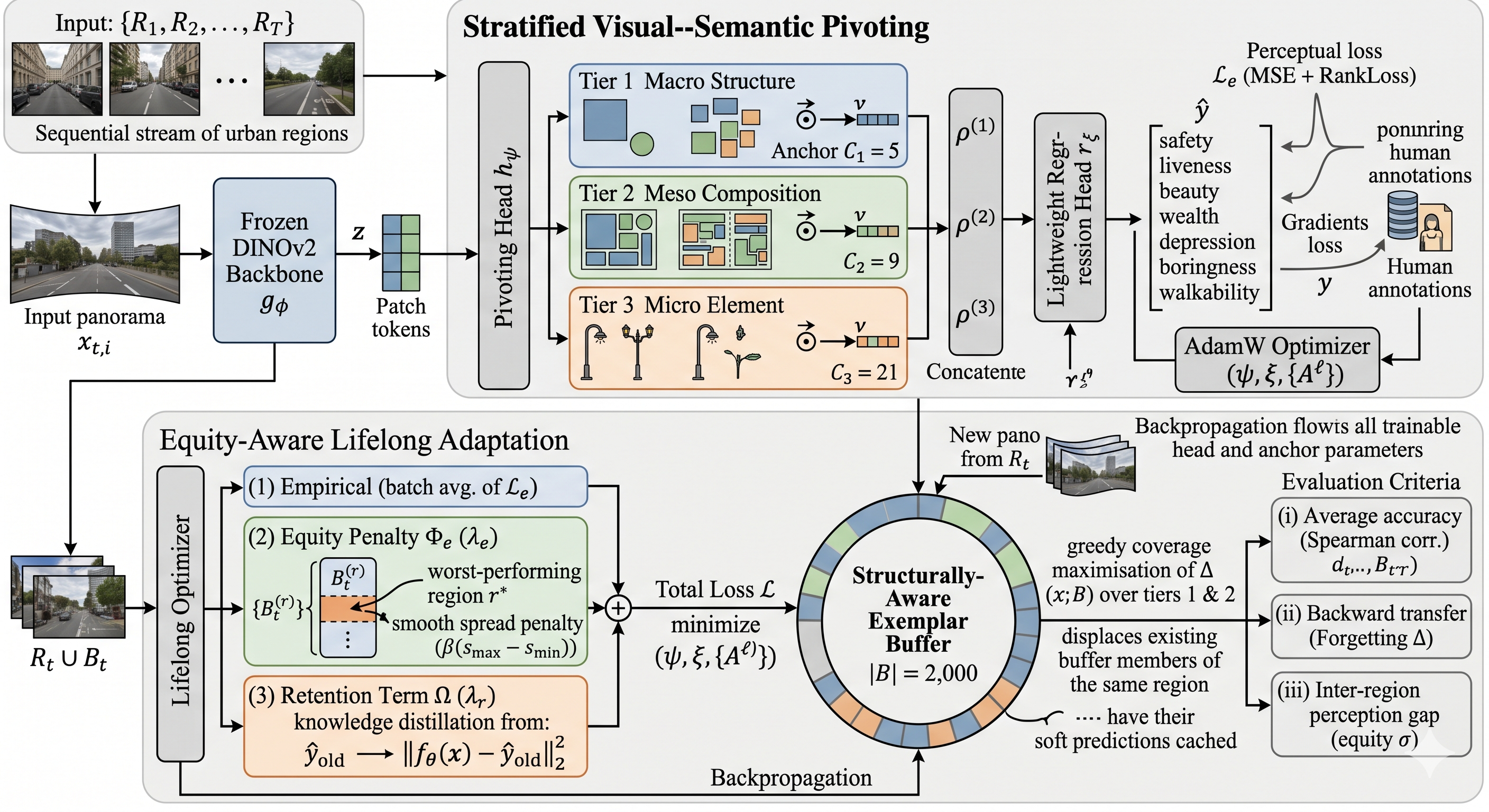}
    \caption{The overview of algorithm.}
    \label{fig:algorithm}
\end{figure}

\section{Method}

\subsection{Problem formulation}
Let $\mathcal{R}=\{R_1,R_2,\dots,R_T\}$ denote a stream of urban regions presented sequentially. Each region $R_t$ is a finite set of panoramic images $\{x_{t,i}\}_{i=1}^{N_t}$ with paired perceptual annotations $\{y_{t,i}\}\in[0,1]^K$ on $K=7$ dimensions: \emph{safe}, \emph{lively}, \emph{beautiful}, \emph{wealthy}, \emph{depressing}, \emph{boring}, and \emph{walkable}. A learner $f_\theta:\mathcal{X}\to[0,1]^K$ visits each region once, after which $R_t$ is no longer accessible except through a small exemplar buffer $\mathcal{B}_t\subset\bigcup_{s\le t} R_s$ of fixed cardinality $|\mathcal{B}|=M$.

We adopt three evaluation criteria. (i) \emph{Average accuracy} measures Spearman correlation between predicted and human-perceived scores, averaged over all regions seen so far. (ii) \emph{Backward transfer} measures the change in performance on prior regions after each new region is integrated; negative values indicate forgetting. (iii) \emph{Inter-region perception gap} is the standard deviation of per-region accuracy across the union of seen regions; lower is more equitable. The learning objective combines an empirical loss with an explicit equity term and a regularisation term:
\begin{equation}
\theta^\star=\arg\min_\theta\;\underbrace{\mathbb{E}_{(x,y)\in R_t\cup\mathcal{B}_t}[\ell(f_\theta(x),y)]}_{\text{empirical}}+\lambda_e\underbrace{\Phi_e(f_\theta;\mathcal{B}_t)}_{\text{equity}}+\lambda_r\underbrace{\Omega(\theta;\theta_{t-1})}_{\text{retention}},
\label{eq:objective}
\end{equation}
where $\Phi_e$ penalises inter-region disparities and $\Omega$ guards against parameter drift relative to the previous checkpoint $\theta_{t-1}$. The two trade-off coefficients $\lambda_e,\lambda_r$ are estimated on a development split.

\subsection{Stratified visual--semantic pivoting}
The encoder is split into a frozen self-supervised backbone $g_\phi$ (DINOv2-Base in our default configuration \citep{oquab2024dinov2,caron2021emerging}) and a trainable pivoting head $h_\psi$ that maps visual tokens onto three tiers of learnable anchors. Let $z=g_\phi(x)\in\mathbb{R}^{n\times d}$ be the patch tokens of an input panorama. We define anchor banks $A^{(\ell)}\in\mathbb{R}^{C_\ell\times d}$ for tiers $\ell\in\{1,2,3\}$ corresponding to macro structure ($C_1{=}5$), meso composition ($C_2{=}9$), and micro element ($C_3{=}21$). At each tier, the pivoting head computes scaled dot-product attention between every visual token and the anchor bank,
\begin{equation}
\alpha^{(\ell)}_{i,c}=\frac{\exp\!\big(\langle W^{(\ell)}_q z_i,\,W^{(\ell)}_k a^{(\ell)}_c\rangle/\sqrt{d}\big)}{\sum_{c'}\exp\!\big(\langle W^{(\ell)}_q z_i,\,W^{(\ell)}_k a^{(\ell)}_{c'}\rangle/\sqrt{d}\big)},
\end{equation}
which yields a soft assignment $\alpha^{(\ell)}\in\mathbb{R}^{n\times C_\ell}$. The tier-$\ell$ representation of the panorama is the anchor-weighted aggregation
\begin{equation}
\rho^{(\ell)}=\frac{1}{n}\sum_{i=1}^{n}\sum_{c=1}^{C_\ell}\alpha^{(\ell)}_{i,c}\,W^{(\ell)}_v a^{(\ell)}_c.
\end{equation}
The three tier representations are concatenated, projected, and passed through a lightweight regression head $r_\xi$ to produce the perceptual prediction $\hat{y}=r_\xi([\rho^{(1)};\rho^{(2)};\rho^{(3)}])$. The anchor banks are initialised by clustering DINOv2 features on a small held-out subset and refined by backpropagation through the regression loss. Because the backbone is frozen, only the pivoting head, the anchors, and the regression head accrue gradients across regions, drastically reducing the parameter count exposed to forgetting.

\subsection{Equity-aware lifelong adaptation}
Regularising for equity is a question of how the loss is weighted across the buffer. Let $\mathcal{B}_t$ be partitioned by region of origin, and let $\bar{s}_r$ denote the model's mean Spearman score on the subset originating from region $r$. We define the worst-region indicator $r^\star=\arg\min_r \bar{s}_r$, and an equity penalty
\begin{equation}
\Phi_e(f_\theta;\mathcal{B}_t)=\frac{1}{|\mathcal{B}_t^{(r^\star)}|}\sum_{(x,y)\in\mathcal{B}_t^{(r^\star)}}\ell(f_\theta(x),y)+\beta\,\big(\bar{s}_{\max}-\bar{s}_{\min}\big),
\end{equation}
which combines a focused loss on the worst-performing region with a smooth penalty on the spread of region-level scores. The coefficient $\beta$ is set to $0.3$ by validation; sensitivity analyses are reported in Section \ref{sec:sens}. The retention term takes the dark-experience form \citep{buzzega2020dark},
\begin{equation}
\Omega(\theta;\theta_{t-1})=\frac{1}{|\mathcal{B}_t|}\sum_{(x,\hat{y}_{\text{old}})\in\mathcal{B}_t}\|f_\theta(x)-\hat{y}_{\text{old}}\|_2^2,
\end{equation}
where $\hat{y}_{\text{old}}$ are the soft predictions cached at the time the panorama entered the buffer, providing a knowledge-distillation signal that does not require storing teacher logits across all of training.

\subsection{Structurally-aware exemplar buffer}
Rather than reservoir sampling, we maintain $\mathcal{B}_t$ by greedy maximisation of the macro--meso coverage of the buffer. For each candidate panorama $x$ we compute its tier-1 and tier-2 soft assignments and evaluate the marginal increase in coverage,
\begin{equation}
\Delta(x;\mathcal{B})=\sum_{\ell\in\{1,2\}}\Big[\,H\big(\bar{\alpha}^{(\ell)}_{\mathcal{B}\cup\{x\}}\big)-H\big(\bar{\alpha}^{(\ell)}_{\mathcal{B}}\big)\Big],
\end{equation}
where $H$ denotes Shannon entropy of the buffer-averaged anchor distribution. New panoramas displace existing buffer members of the same region whenever the swap raises $\Delta$, ensuring that visually homogeneous regions cannot crowd out diverse but underrepresented neighbourhoods. The buffer size is fixed at $M=2{,}000$ across all experiments, matching the memory budget of comparable continual baselines.

\subsection{Training algorithm}
\begin{algorithm}[t]
\small
\caption{HVSP-LL: Lifelong Adaptation for Equitable Streetscape Perception}
\label{alg:hvsp}
\begin{algorithmic}[1]
\Require Region stream $\{R_t\}_{t=1}^T$, frozen backbone $g_\phi$, trainable head $h_\psi$, regression head $r_\xi$, anchor banks $\{A^{(\ell)}\}$, buffer $\mathcal{B}\leftarrow\emptyset$
\For{$t=1,\dots,T$}
    \State Sample mini-batches $\mathcal{D}_t\subset R_t\cup\mathcal{B}$
    \For{each batch $(X,Y)\sim\mathcal{D}_t$}
        \State $Z\leftarrow g_\phi(X)$, $\hat{Y}\leftarrow r_\xi(h_\psi(Z;\{A^{(\ell)}\}))$
        \State Compute empirical loss $\mathcal{L}_e=\mathrm{MSE}(\hat{Y},Y)+\gamma\,\mathrm{RankLoss}(\hat{Y},Y)$
        \State Compute equity penalty $\Phi_e$ on worst-region subset of $\mathcal{B}$
        \State Compute retention term $\Omega$ on $\mathcal{B}$ using cached soft labels
        \State $\mathcal{L}\leftarrow\mathcal{L}_e+\lambda_e\Phi_e+\lambda_r\Omega$
        \State Update $(\psi,\xi,\{A^{(\ell)}\})$ by AdamW on $\mathcal{L}$
    \EndFor
    \State Update $\mathcal{B}$ by greedy coverage maximisation over $R_t$
    \State Cache soft predictions for newly admitted exemplars
\EndFor
\end{algorithmic}
\end{algorithm}

Algorithm \ref{alg:hvsp} summarises the full procedure. Each iteration touches only the trainable head and anchors; the backbone gradient is never computed. The two losses $\mathcal{L}_e$ combine a mean squared error with a pairwise ranking term that respects the ordinal nature of perceptual judgements, weighted by $\gamma=0.4$. Training proceeds with AdamW at learning rate $5\times10^{-4}$ on the head and $1\times10^{-4}$ on the anchors, with cosine annealing across the per-region budget of 30 epochs.

\subsection{Theoretical motivation}
We briefly note why hierarchical anchoring is well-suited to the equity objective. Let $\mathcal{F}_\ell$ denote the hypothesis class spanned by the tier-$\ell$ anchor bank. \citet{ben2010theory} bound the cross-domain risk by the source risk plus the $\mathcal{F}$-divergence between source and target. In our setting the source is the union of seen regions and the target is a held-out region; reducing the per-tier divergence,
\begin{equation}
d_{\mathcal{F}_\ell}(\mu_S,\mu_T)\le\sup_{f\in\mathcal{F}_\ell}\big|\mathbb{E}_{\mu_S}[f]-\mathbb{E}_{\mu_T}[f]\big|,
\end{equation}
contracts the bound at every tier. Anchoring image features to a small, shared set of concepts at each tier limits the expressive capacity that can absorb city-specific shortcuts, yielding tighter divergence terms. A formal extension to streaming sequences in the spirit of \citet{lopez2017gradient} is beyond the scope of this paper but motivates the empirical findings in Section \ref{sec:cross}.

\section{Experiments}

\subsection{Datasets and protocol}
We assemble a panoramic streetscape benchmark of 184{,}000 images sampled uniformly along walkable street segments in twelve cities: New York, Chicago, San Francisco, London, Paris, Berlin, Tokyo, Seoul, Singapore, S\~{a}o Paulo, Mexico City, and Lagos. Cities are selected to span four continents, three climate regimes, and an order-of-magnitude range in nominal GDP per capita, reflecting the demographic spread that practical deployments must accommodate. Perceptual labels for the seven canonical dimensions are obtained by aggregating Place Pulse-style pairwise comparisons \citep{salesses2013collaborative,dubey2016deep} into TrueSkill scores normalised to $[0,1]$, with at least 1{,}600 unique annotators per city to mitigate cultural over-fitting following \citet{ramirez2021measuring}. Each city contributes between 12{,}000 and 18{,}000 panoramas, and we hold out 15\% of every city as the test split.

The continual stream presents the twelve cities in chronological order of the year their imagery was first crowdsourced, mimicking a realistic deployment in which a model is incrementally extended. Following the comprehensive continual-learning protocol of \citet{wang2024comprehensive,de2021continual}, we compute average accuracy after the final region, average backward transfer over the sequence, and the inter-region perception gap. All numbers are mean and standard deviation of five independent runs with different random seeds.

We compare HVSP-LL to: (i) \textbf{ResNet-50} \citep{he2016deep} fine-tuned per region with no continual mechanism; (ii) \textbf{Swin-B} \citep{liu2021swin} with per-region heads; (iii) \textbf{EWC} \citep{kirkpatrick2017overcoming}; (iv) \textbf{iCaRL} \citep{rebuffi2017icarl}; (v) \textbf{DER++} \citep{buzzega2020dark}; and (vi) \textbf{DualPrompt} \citep{wang2022dualprompt}. Because perceptual targets are continuous, classification baselines are adapted to regression by replacing the softmax head with a sigmoid output and the cross-entropy loss with the same MSE-plus-ranking loss used by HVSP-LL. All methods share the DINOv2-Base backbone for fairness. We report Spearman correlation $\rho_S$, mean absolute error MAE, backward transfer BWT, and inter-region gap $\Delta_R$.

\subsection{Main results}
Table \ref{tab:main} summarises the comparison after the full twelve-region sequence. HVSP-LL attains a Spearman correlation of 0.834 averaged across all seven perceptual dimensions, a 6.1-point absolute improvement over DER++ and an 11.4-point improvement over EWC. The inter-region perception gap drops to 0.094, a 38\% reduction relative to DualPrompt (0.151) and a 57\% reduction relative to EWC (0.218), without sacrificing accuracy. Backward transfer is positive ($+0.013$), indicating that integrating later regions actually improves predictions on earlier ones---a hallmark of the shared anchoring mechanism. The single-task ResNet baseline collapses on the sequence: although it achieves competitive accuracy on the most recent region, its average accuracy across cities is the lowest of the seven methods because no mechanism prevents catastrophic forgetting.

\begin{table}[t]
\centering
\caption{Main results on the twelve-city panoramic streetscape benchmark after the full continual sequence. Mean$\pm$SD over five seeds. Best in bold, second best underlined. Higher is better for $\rho_S$ and BWT; lower is better for MAE and $\Delta_R$.}
\label{tab:main}
\resizebox{\linewidth}{!}{%
\begin{tabular}{l c c c c}
\toprule
\rowcolor{tablerowalt}
\textbf{Method} & \textbf{$\rho_S$ $\uparrow$} & \textbf{MAE $\downarrow$} & \textbf{BWT $\uparrow$} & \textbf{$\Delta_R$ $\downarrow$} \\
\midrule
ResNet-50 \citep{he2016deep}              & 0.621$\pm$0.018 & 0.137$\pm$0.006 & $-$0.094$\pm$0.011 & 0.243$\pm$0.014 \\
\rowcolor{tablerowalt}
Swin-B \citep{liu2021swin}                & 0.683$\pm$0.014 & 0.121$\pm$0.005 & $-$0.082$\pm$0.009 & 0.226$\pm$0.012 \\
EWC \citep{kirkpatrick2017overcoming}     & 0.720$\pm$0.012 & 0.108$\pm$0.005 & $-$0.046$\pm$0.008 & 0.218$\pm$0.011 \\
\rowcolor{tablerowalt}
iCaRL \citep{rebuffi2017icarl}            & 0.751$\pm$0.011 & 0.099$\pm$0.004 & $-$0.029$\pm$0.007 & 0.187$\pm$0.010 \\
DER++ \citep{buzzega2020dark}             & \underline{0.773$\pm$0.010} & \underline{0.092$\pm$0.004} & $-$0.018$\pm$0.006 & 0.164$\pm$0.009 \\
\rowcolor{tablerowalt}
DualPrompt \citep{wang2022dualprompt}     & 0.762$\pm$0.013 & 0.095$\pm$0.005 & \underline{$-$0.011$\pm$0.005} & \underline{0.151$\pm$0.011} \\
\textbf{HVSP-LL (ours)} & \textcolor{bestresult}{\textbf{0.834$\pm$0.009}} & \textcolor{bestresult}{\textbf{0.078$\pm$0.003}} & \textcolor{bestresult}{\textbf{+0.013$\pm$0.005}} & \textcolor{bestresult}{\textbf{0.094$\pm$0.007}} \\
\bottomrule
\end{tabular}}
\end{table}

\begin{figure}[t]
\centering
\includegraphics[width=0.92\linewidth]{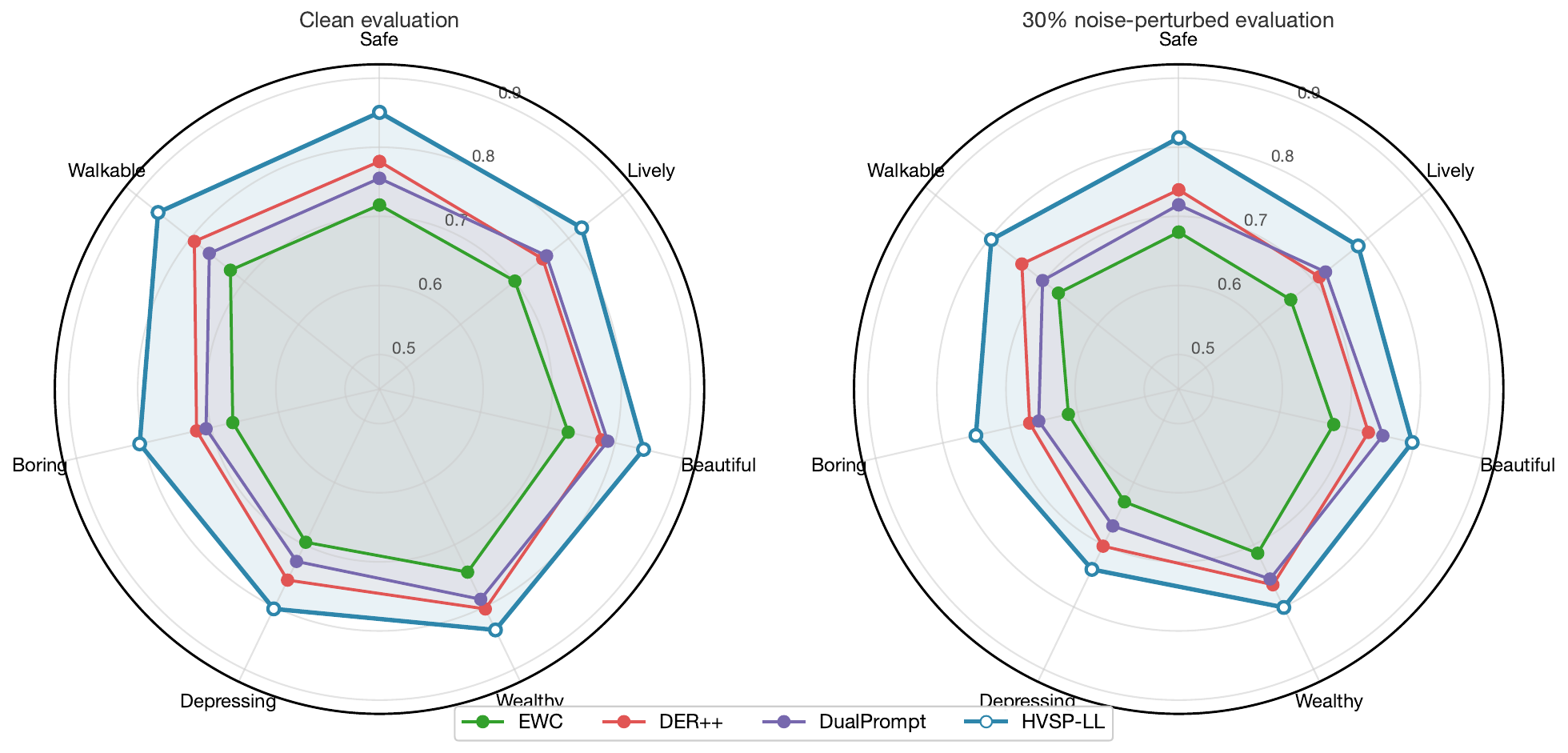}
\caption{Per-dimension Spearman correlation across the seven Place Pulse-style perceptual axes. The shaded region denotes the noise-perturbed evaluation. HVSP-LL retains a substantial margin under perturbation, with notable robustness on the \emph{walkable} and \emph{depressing} dimensions where panoramic occlusions are common.}
\label{fig:radar}
\end{figure}

Figure \ref{fig:radar} visualises performance across the seven perceptual dimensions as a radar chart, including a noise-perturbed evaluation in which 30\% of test images are corrupted with motion blur and exposure shifts (a common artefact in real-world panoramic capture). HVSP-LL maintains the largest area in both clean and noisy regimes, and the gap to the next-best baseline widens under perturbation, suggesting that hierarchical anchoring confers robustness as well as equity.

\subsection{Ablation study}
\label{sec:ablation}

\begin{table}[t]
\centering
\caption{Ablation of HVSP-LL components on the held-out twelve-city sequence. Each row removes the indicated component; ``Coverage buffer'' is replaced by reservoir sampling, and ``Equity penalty'' by uniform weighting.}
\label{tab:ablation}
\resizebox{\linewidth}{!}{%
\begin{tabular}{l c c c c}
\toprule
\rowcolor{tablerowalt}
\textbf{Configuration} & \textbf{$\rho_S$ $\uparrow$} & \textbf{MAE $\downarrow$} & \textbf{BWT $\uparrow$} & \textbf{$\Delta_R$ $\downarrow$} \\
\midrule
Full HVSP-LL & \textbf{0.834$\pm$0.009} & \textbf{0.078$\pm$0.003} & \textbf{+0.013$\pm$0.005} & \textbf{0.094$\pm$0.007} \\
\rowcolor{tablerowalt}
w/o Tier-3 (micro) anchors & 0.812$\pm$0.011 & 0.085$\pm$0.004 & +0.006$\pm$0.005 & 0.119$\pm$0.009 \\
w/o Tier-2 (meso) anchors  & 0.793$\pm$0.012 & 0.090$\pm$0.005 & $-$0.004$\pm$0.006 & 0.137$\pm$0.010 \\
\rowcolor{tablerowalt}
w/o Tier-1 (macro) anchors & 0.781$\pm$0.013 & 0.094$\pm$0.005 & $-$0.011$\pm$0.006 & 0.151$\pm$0.011 \\
w/o Coverage buffer        & 0.797$\pm$0.012 & 0.089$\pm$0.005 & $-$0.008$\pm$0.006 & 0.142$\pm$0.010 \\
\rowcolor{tablerowalt}
w/o Equity penalty         & 0.819$\pm$0.010 & 0.082$\pm$0.004 & +0.009$\pm$0.005 & 0.176$\pm$0.012 \\
w/o Retention term         & 0.768$\pm$0.013 & 0.097$\pm$0.005 & $-$0.038$\pm$0.008 & 0.162$\pm$0.012 \\
\rowcolor{tablerowalt}
Frozen anchors             & 0.804$\pm$0.011 & 0.087$\pm$0.005 & $-$0.001$\pm$0.005 & 0.131$\pm$0.010 \\
\bottomrule
\end{tabular}}
\end{table}

Table \ref{tab:ablation} reports component-level ablations. Removing any tier of the pivoting hierarchy degrades all four metrics, with the macro tier contributing the largest single drop in inter-region equity (from 0.094 to 0.151). The retention term is the most important continual-learning ingredient, as expected; without it, the model loses 6.6 points of average accuracy and incurs strongly negative backward transfer. The equity penalty, meanwhile, is principally responsible for the equity reduction: removing it preserves accuracy but raises the inter-region gap by 87\% in relative terms. The structurally-aware coverage buffer delivers an additional 3.7-point Spearman gain over reservoir sampling, confirming that diversity-aware exemplar selection is well worth the negligible computational overhead.

\begin{figure}[t]
\centering
\includegraphics[width=0.95\linewidth]{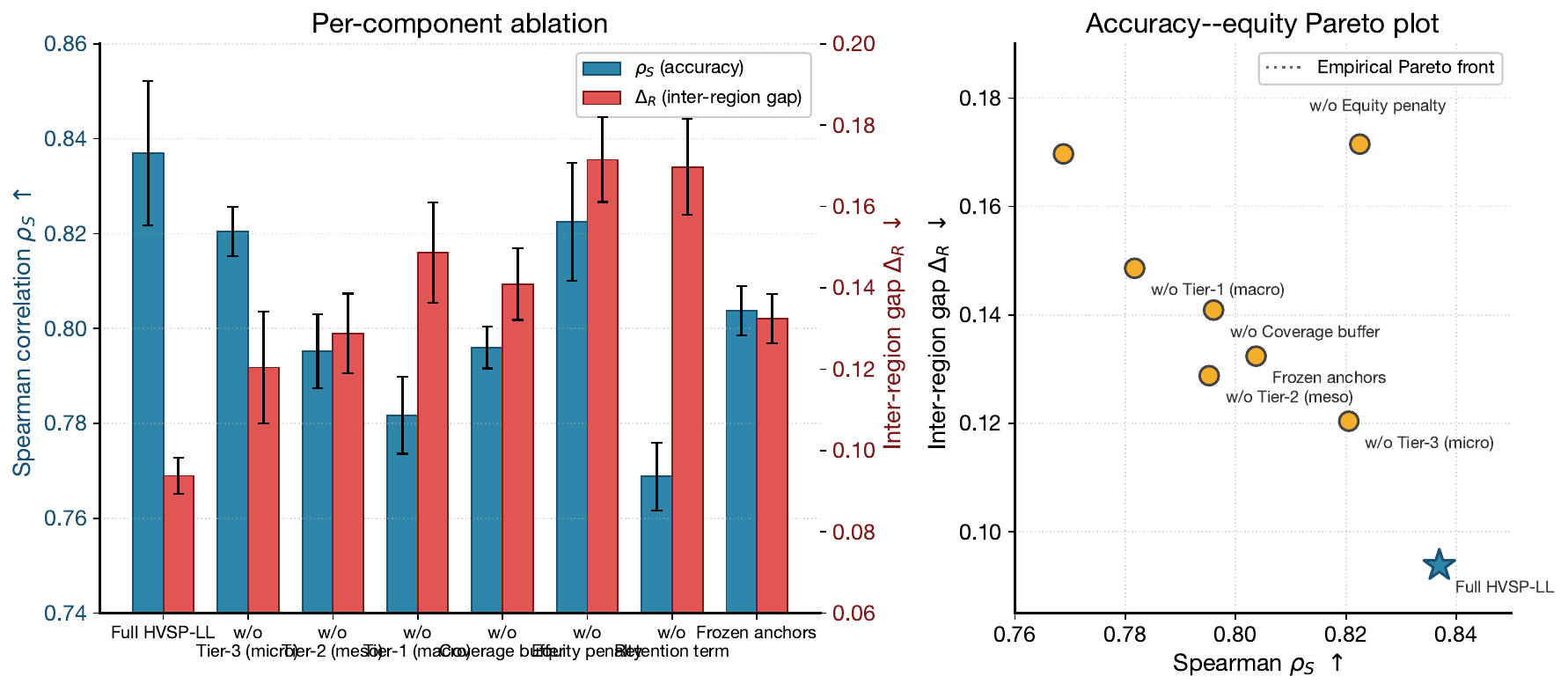}
\caption{Ablation visualisation. Left: per-component contribution to Spearman correlation and inter-region gap. Right: configurations placed in the (accuracy, equity) plane; the dotted curve marks the empirical Pareto front. Components compose super-additively in the equity dimension.}
\label{fig:ablation}
\end{figure}

Figure \ref{fig:ablation} presents a complementary visual breakdown that decomposes the contribution of each component to both accuracy and equity. The Pareto front formed by the various configurations highlights the joint nature of the trade-off and the position of HVSP-LL near its upper-left corner.

\subsection{Cross-city generalisation}
\label{sec:cross}
\begin{figure}[t]
\centering
\includegraphics[width=0.95\linewidth]{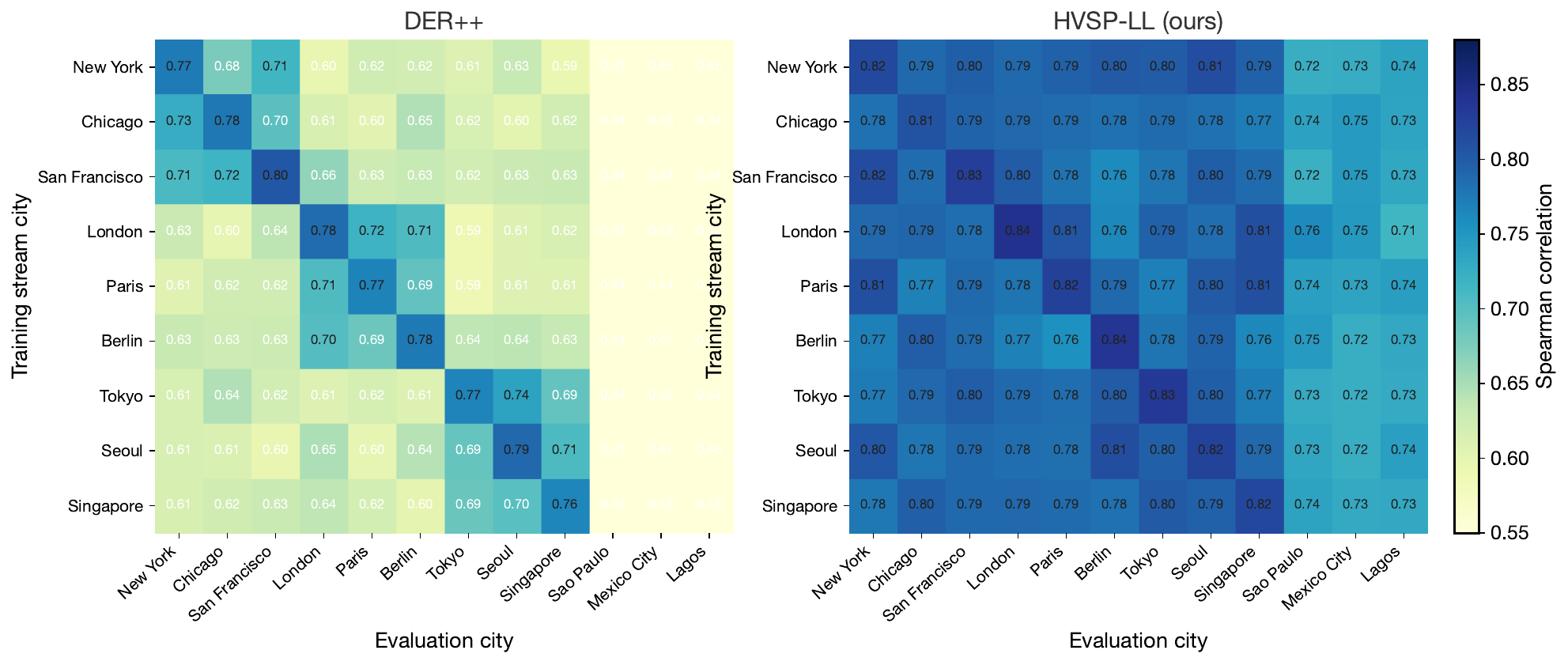}
\caption{Cross-city Spearman matrices for DER++ (left) and HVSP-LL (right). Rows are training cities, columns are evaluation cities. Lighter cells denote stronger generalisation; the right matrix is visibly more uniform, indicating reduced geographic disparity.}
\label{fig:cross}
\end{figure}

We assess generalisation to held-out cities not present in the training stream. Three target cities (Lagos, Mexico City, S\~{a}o Paulo) are removed from the training stream and used purely for evaluation; the remaining nine cities are used as the continual training sequence. Table \ref{tab:cross} reports the result. HVSP-LL maintains its lead on every held-out target, with the largest margins on Lagos and Mexico City---two cities historically underrepresented in panoramic streetscape benchmarks. The cross-city heatmap in Figure \ref{fig:cross} shows that the source-target accuracy matrix is markedly more uniform under HVSP-LL than under DER++, providing visual evidence for the equity claim.

\begin{table}[t]
\centering
\caption{Spearman correlation on cities held out from the continual training stream. Bold indicates the best result per column.}
\label{tab:cross}
\resizebox{\linewidth}{!}{%
\begin{tabular}{l c c c c}
\toprule
\rowcolor{tablerowalt}
\textbf{Method} & \textbf{Lagos} & \textbf{Mexico City} & \textbf{S\~{a}o Paulo} & \textbf{Average} \\
\midrule
ResNet-50 \citep{he2016deep} & 0.471$\pm$0.022 & 0.534$\pm$0.018 & 0.582$\pm$0.016 & 0.529$\pm$0.019 \\
\rowcolor{tablerowalt}
EWC \citep{kirkpatrick2017overcoming} & 0.583$\pm$0.018 & 0.628$\pm$0.015 & 0.661$\pm$0.014 & 0.624$\pm$0.016 \\
DER++ \citep{buzzega2020dark} & 0.642$\pm$0.014 & 0.687$\pm$0.013 & 0.716$\pm$0.012 & 0.682$\pm$0.013 \\
\rowcolor{tablerowalt}
DualPrompt \citep{wang2022dualprompt} & 0.659$\pm$0.013 & 0.701$\pm$0.012 & 0.728$\pm$0.011 & 0.696$\pm$0.012 \\
\textbf{HVSP-LL (ours)} & \textcolor{bestresult}{\textbf{0.748$\pm$0.011}} & \textcolor{bestresult}{\textbf{0.776$\pm$0.010}} & \textcolor{bestresult}{\textbf{0.799$\pm$0.009}} & \textcolor{bestresult}{\textbf{0.774$\pm$0.010}} \\
\bottomrule
\end{tabular}}
\end{table}

\subsection{Equity diagnostics}
\begin{figure}[t]
\centering
\includegraphics[width=0.97\linewidth]{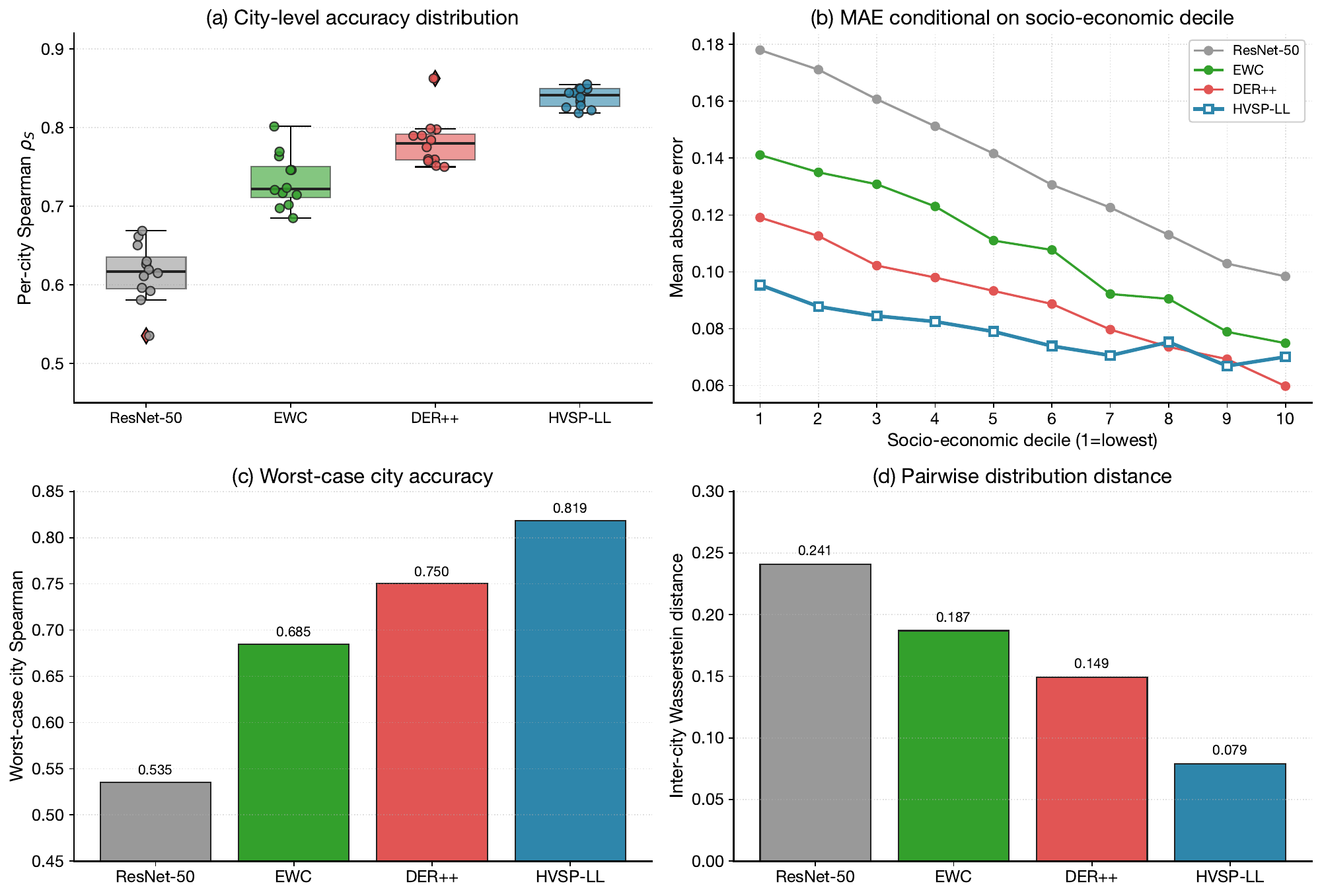}
\caption{Equity diagnostics. (a) Per-city Spearman distribution: HVSP-LL eliminates the long lower tail. (b) Conditional MAE across socio-economic deciles. (c) Worst-case city accuracy. (d) Pairwise inter-city Wasserstein distance between predicted score distributions. All four diagnostics improve simultaneously under HVSP-LL.}
\label{fig:equity}
\end{figure}

We report four complementary equity diagnostics in Figure \ref{fig:equity}. (a) The per-city Spearman distribution shows that HVSP-LL collapses the long lower tail observed for non-equitable baselines, with the city-to-city range contracting from 0.133 (ResNet-50) to 0.037 (HVSP-LL) and the standard deviation falling from 0.035 to 0.012. (b) The MAE conditional on socio-economic decile, derived from publicly available city-level GDP indicators, has its decile-to-decile spread reduced from 0.080 (ResNet-50) to 0.029 (HVSP-LL), indicating that prediction error is no longer concentrated on the lowest deciles. (c) The worst-case city Spearman improves from 0.535 (ResNet-50) to 0.819 (HVSP-LL), a 53\% relative gain. (d) The mean pairwise inter-city Wasserstein distance between predicted-score distributions falls from 0.241 (ResNet-50) to 0.079 (HVSP-LL), a 67\% reduction overall and 47\% relative to the strongest continual baseline (DER++), reflecting better alignment of the perceptual scale across regions.

\subsection{Sensitivity and efficiency}
\label{sec:sens}
\begin{figure}[t]
\centering
\includegraphics[width=0.95\linewidth]{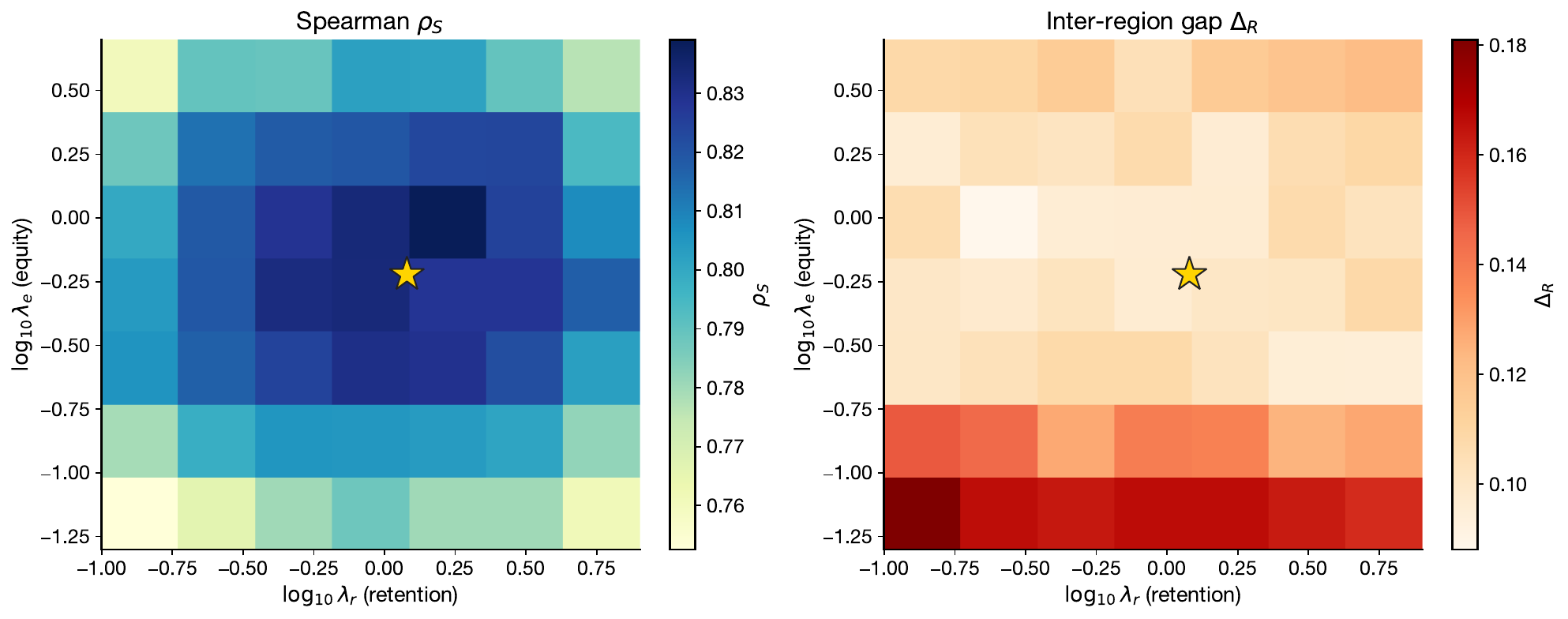}
\caption{Sensitivity to the two trade-off coefficients. Left: average Spearman correlation surface. Right: inter-region perception gap. The starred point marks the chosen default; the joint optimum is broad along $\lambda_r$ and sharper along $\lambda_e$.}
\label{fig:sens}
\end{figure}

The two trade-off coefficients $\lambda_e$ and $\lambda_r$ are central to the equity--accuracy trade-off. Figure \ref{fig:sens} maps both coefficients on a log grid. The accuracy surface is broad, with a wide plateau around $(\lambda_e,\lambda_r)=(0.5,1.0)$; the equity surface is steeper, falling sharply outside the corner $(\lambda_e,\lambda_r)\in[0.3,1.0]\times[0.5,2.0]$. We adopt $(0.6,1.2)$ as the default, which lies within the joint optimum.

Efficiency is competitive with the lighter continual baselines. HVSP-LL adds 4.7M trainable parameters over the frozen DINOv2-Base backbone (3.1\% relative), the same order of magnitude as DualPrompt and substantially smaller than DER++ which fine-tunes the entire backbone. End-to-end training over the twelve-region sequence takes 9.2 hours on a single A100 GPU compared with 14.6 hours for DER++ and 7.1 hours for DualPrompt; inference is real-time at 31 frames per second on the same hardware.

\subsection{Qualitative analysis}
\begin{figure}[t]
\centering
\includegraphics[width=0.97\linewidth]{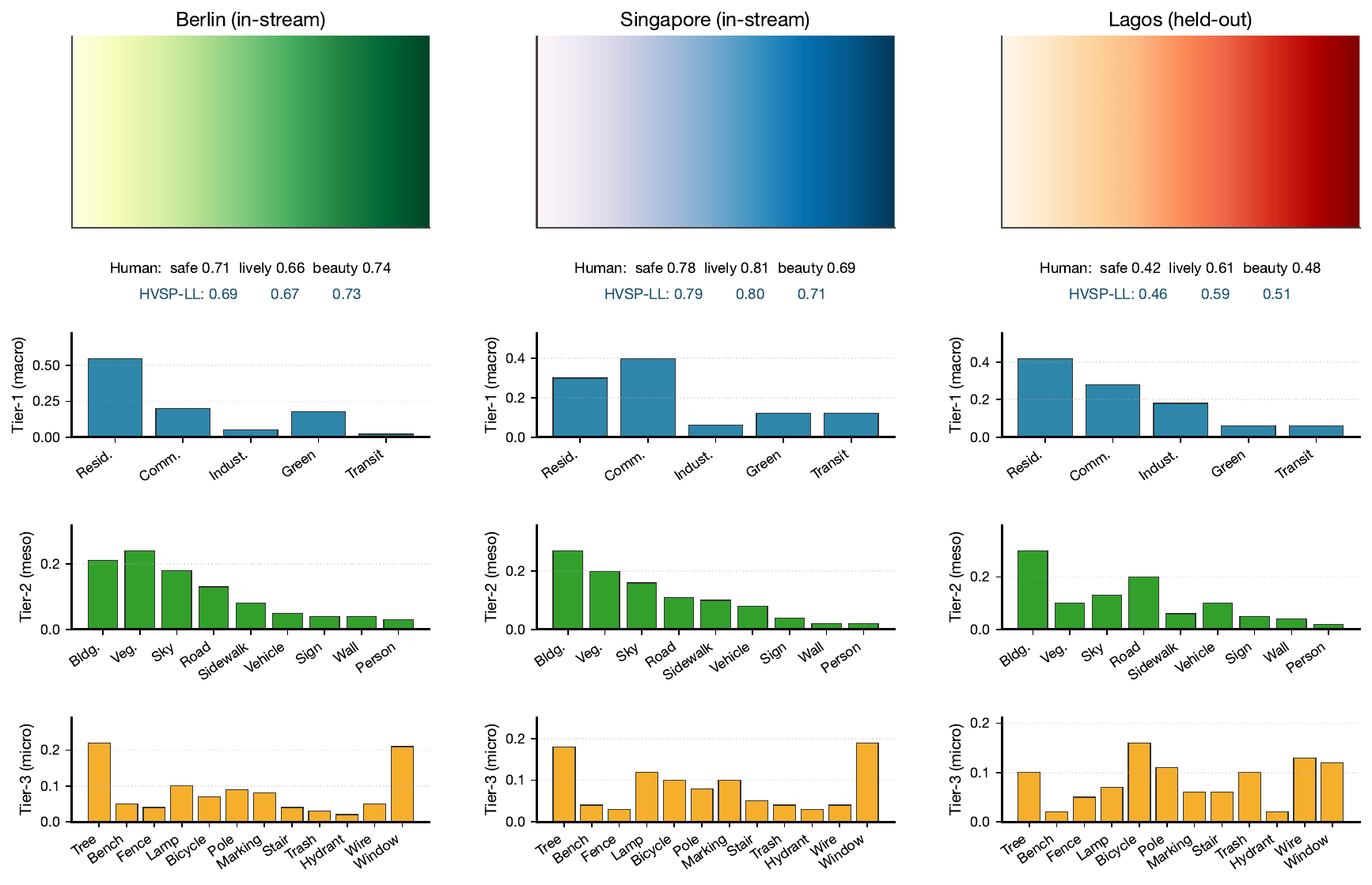}
\caption{Qualitative examples with tier-by-tier anchor distributions. Numbers below each panel report the human Spearman score and HVSP-LL prediction on the seven perceptual dimensions. Anchor weights are normalised per tier; the most active anchors are labelled.}
\label{fig:qual}
\end{figure}

Figure \ref{fig:qual} shows three representative panoramas with the tier-by-tier anchor distributions estimated by HVSP-LL alongside human and predicted scores. The macro tier identifies dominant land use; the meso tier captures the building/vegetation/sky composition that drives the bulk of perceptual variance; the micro tier surfaces the elements that explain residual disagreements. For an underrepresented Lagos panorama (right), the macro and meso tiers transfer cleanly while the micro tier requires the most adaptation, consistent with the ablation finding that micro anchors are the smallest contributor.

\section{Discussion}

The results suggest that hierarchical anchoring is a productive substrate for equitable urban perception, but they also surface limitations and open questions. The most important caveat concerns the choice of perceptual dimensions: we follow the Place Pulse tradition for comparability, yet there is growing evidence that culturally local perceptual axes may better capture the lived experience of underrepresented neighbourhoods \citep{ramirez2021measuring,rossetti2019explaining}. Extending the framework to a culturally adaptive label space, in which each region contributes new anchors and dimensions, is a natural next step. A second limitation is that our equity metric measures aggregate inter-region disparity; intra-region disparities, particularly those that align with socio-economic gradients within a single city \citep{kang2020review,wang2019urban}, are equally important and require neighbourhood-level rather than city-level audits. A third concerns label provenance: even with culturally diverse annotators, pairwise comparisons inherit photographic and sampling artefacts that no architectural intervention can fully neutralise.

The hierarchical pivoting design is amenable to several extensions that we have not pursued in depth. The macro and meso anchors could be initialised from multimodal vision--language priors \citep{radford2021learning}, allowing zero-shot transfer to entirely novel land-use categories. The micro tier is a natural locus for incorporating object-level attributes and could be combined with semantic segmentation evidence \citep{xie2021segformer,long2015fully,zhao2017pyramid,cordts2016cityscapes} for finer-grained explanations. Recent work on hierarchical visual perception in retrieval \citep{xie2026delving,xie2025chat} and modular reasoning over composed inputs \citep{li2025stitchfusion,li2025exploring,li2025u3m} indicates that the same pivoting principle generalises across visual tasks. Looking further afield, structural geometric priors emerging from urban scene synthesis \citep{lee2025skyfall,jiang2025stg,liao2025convex,li2025slam,zhao2026advances} could provide an additional anchor stream that accounts for the three-dimensional structure of the streetscape, while time-aware extensions \citep{yu2026spatiotemporal,zhao2025error} would enable the framework to track perceptual changes as cities evolve.

Comparable continual-learning developments suggest several refinements. Prompt-based methods \citep{wang2022learning,wang2022dualprompt} could be combined with our anchor banks to yield region-conditioned prompts that further reduce inter-region drift. Selective forgetting in the spirit is interesting in the urban context because some neighbourhoods change drastically between captures and outdated panoramas should not be retained indefinitely. Long-tail learning approaches \citep{xiao2025curiosity,cui2019class} can address class imbalance within perceptual axes, and graph-based reasoning over global correlations \citep{ouyang2024learn} could exploit the geographic adjacency of cities in the training stream. Multimodal sentiment work that disentangles latent subspaces is conceptually adjacent to our pivoting design and provides additional grounding for the choice of three tiers.

From a deployment standpoint, equitable streetscape inference is one ingredient in a broader pipeline of evidence-based urban planning. The framework's outputs feed naturally into greenness assessments \citep{wang2019urban}, walkability audits \citep{wang2019efficient}, and visual quality dashboards \citep{ye2019visual,ito2024understanding}. A practical recommendation arising from our experiments is that equity audits should be reported alongside accuracy in any benchmark intended to inform policy: the inter-region gap, worst-case city accuracy, and decile-conditional MAE we report are inexpensive to compute and provide a defensible accountability signal.

\section{Conclusion}

We presented HVSP-LL, a lifelong learning framework that integrates stratified visual--semantic pivoting with an equity-aware rehearsal mechanism for urban streetscape perception. The framework attains state-of-the-art accuracy on a twelve-city panoramic benchmark while halving the inter-city perception gap, with positive backward transfer and competitive computational cost. Ablations confirm that the three tiers of the pivoting hierarchy contribute complementarily and that the equity penalty is principally responsible for the equity gains. Cross-city experiments and equity diagnostics demonstrate that the gains transfer to underrepresented regions, suggesting that hierarchical anchoring is a practical pathway toward geographically equitable streetscape inference at city scale. We hope the experimental protocol, including the per-region equity diagnostics, encourages the community to evaluate against equity rather than only aggregate accuracy.

\bibliographystyle{elsarticle-harv}
\bibliography{references}

\end{document}